# Ultralow-dimensionality reduction for identifying critical transitions by spatial-temporal PCA

Pei Chen[1], Yaofang Suo[1], Rui Liu[1,*], Luonan Chen[2,*]

[1] School of Mathematics, South China University of Technology, Guangzhou 510640, China.

[2] Key Laboratory of Systems Biology, Shanghai Institute of Biochemistry and Cell Biology, Chinese Academy of Sciences, Shanghai 200031, China.

*Correspondence: Luonan Chen, lnchen@sibs.ac.cn; Rui Liu, scliurui@scut.edu.cn

## Abstract

Discovering dominant patterns and exploring dynamic behaviors especially critical state transitions and tipping points in high-dimensional time-series data are challenging tasks in study of real-world complex systems, which demand interpretable data representations to facilitate comprehension of both spatial and temporal information within the original data space. Here, we proposed a general and analytical ultralow-dimensionality reduction method for dynamical systems named spatial-temporal principal component analysis (stPCA) to fully represent the dynamics of a high-dimensional time-series by only a single latent variable without distortion, which transforms high-dimensional spatial information into one-dimensional temporal information based on nonlinear delay-embedding theory. The dynamics of this single variable is analytically solved and theoretically preserves the temporal property of original high-dimensional time-series, thereby accurately and reliably identifying the tipping point before an upcoming critical transition. Its applications to real-world datasets such as individual-specific heterogeneous ICU records demonstrated the effectiveness of stPCA, which quantitatively and robustly provides the early-warning signals of the critical/tipping state on each patient.

## Teaser

The stPCA fully represents the dynamics of high-dimensional time series by a single latent variable, and can effectively identify critical transitions.

## Introduction

Numerous physical and biological processes are characterized by high-dimensional nonlinear dynamical systems, in which critical state transitions often occur (*1*). However, for most real-world systems which are too complex to be simply described by an explicit model, we have to study their dynamics especially identifying the tipping point before an upcoming critical transition through high-dimensional time-series or sequential data. The exploratory analysis of these systems requires data dimensionality reduction and linear or nonlinear data representations (*2*), which are critical in dynamical analysis, pattern recognition, and visualization (*3-5*). It is still a challenging problem to efficiently perform dimensionality reduction and explore ultralow-dimensional dynamics in an interpretable way for complex dynamical systems. A variety of data-driven techniques have been developed to facilitate this task, including t-distributed stochastic neighbor embedding (t-SNE) (*6*), uniform manifold approximation and projection (UMAP) (*7*), multidimensional scaling (MDS) (*8*), local linear embedding (LLE) (*9*), isometric feature mapping (ISOMAP) (*10*) . Nevertheless, these approaches generally lack the desirable linear and temporal interpretability, which is essential in enabling a more lucid depiction of location, direction, distance and dynamics in the low-dimensional representation space.

Linear dimensionality reduction methods, including principal component analysis (PCA) (*11*), linear discriminant analysis (LDA) (*12*), sparse dictionary learning (SDL) (*13*), and independent component analysis (ICA) (*14*) offer interpretability by preserving the relations among variables using linear transformations. For example, PCA projects the original data onto a lower-dimensional subspace in a linear form such that the projected latent variables capture the maximum variance in the representation space. However, these existing methods do not leverage the temporal or inherent relations among sequentially observed samples within the original data; hence, explaining the dynamic nature of the time-series data by these methods is difficult. Besides, these methods are sensitive to outliers, and thereby susceptible to noise interference. These limitations are a barrier to the practical usage of data representation for real-world dynamical systems.

Deep neural networks (DNNs) have emerged as powerful tools in sequential data analysis and dynamic modeling, leveraging deep learning techniques to handle nonlinear embedding and intricate patterns in dynamical systems, including recurrent neural networks (RNNs) (*15*) and the variants (*16*, *17*), temporal convolutional networks (TCNs) (*18*), autoencoders (AEs) (*19*), deep belief networks (DBNs) (*20*), transformers (*21*), CEBRA(*22*), LFADS(*23*), CANDyMan(*24*), and MARBLE(*25*). Despite of the state-of-the-art performances of these DNN architectures, traditional RNNs can be computationally intensive and may suffer from issues such as vanishing or exploding gradients, which can affect their ability to capture all relevant information. Addtionally, many DNN architectures still demand demand significant computational and memory resources, which can be limiting in resource-constrained environments or scenarios with strict latency requirements.

Recently, the Koopman operator theory offers an alternative perspective, suggesting that linear superposition may characterize nonlinear dynamics through the infinite-dimensional linear Koopman operator (*26–29*), but obtaining its representations has proven difficult in all but the simplest systems. One approach to find tractable finite-dimensional representations of the Koopman operator is DNN (*30*), however, they still face challenges related to the requirement of large training samples and other computational issues. Dynamic mode decomposition (DMD) (*26–28*) is another commonly employed method for approximating the Koopman operator from time-series data. It decomposes the system dynamics into temporal modes, each of which is associated with a fixed oscillation frequency and decay/growth rate. Nonetheless, DMD may require extensive time-series data, is sensitive to noise, and also involves the computation of pseudo-inverses of linear matrix operator and truncation of spectrum, all potentially impacting accuracy. Another method for representing dynamical systems is sparse identification of nonlinear dynamical systems (SINDY) (*31*), which combines sparsity-promoting techniques and machine learning with nonlinear dynamical systems to discover governing equations, but the accuracy highly depends on the choice of measurement coordinates and the sparsifying function.

Furthermore, these dimensionality reduction methods mentioned above generally require multiple variables or principal components to represent the original high-dimensional data. In other words, multiple representation components are usually mandatory for reconstructing a dynamical system by those methods to avoid information loss, which leads to challenges in efficiently visualizing or quantifying nonlinear phenomena, e.g., the critical slowing down (CSD) effect (*3-5*). In fact, when a dynamical system approaches a bifurcation point from a stable steady state, the space is generically constrained to a center manifold, which typically has codim-1 local bifurcation with a dominant real eigenvalue such as the saddle-node bifurcation; thus, this one-dimensional variable is roughly considered to represent the center manifold near the tipping point. A natural question arises for analyzing the observed high-dimensional time-series data: is it possible to devise an ultralow-dimensionality reduction method with only a single latent variable to fully represent or reconstruct the whole dynamics of the original, yet unknown, high-dimensional system?

Based on the generalized Takens' embedding theory (*32, 33*), the dynamics of a system can be topologically reconstructed from the delay embedding scheme if $L > 2d > 0$, where $d$ is the box-counting dimension of the attractor for the original system and $L$ represents the embedding dimension. By assuming that the steady state of a high-dimensional system is contained in a low-dimensional manifold, which is generally satisfied for dissipative real-world systems, the spatial-temporal information (STI) (*34*) transformation has theoretically been derived from the delay embedding theory. This approach transforms the spatial information of high-dimensional data into the temporal dynamics of any (one) target variable, i.e., a one-dimensional delay embedding of this single variable, thus exploiting not only inter-sample (cross-sample) information but also intra-sample information embedded in the high-dimensional/spatial data. Several methods, such as the randomly distributed embedding (RDE) (*34*), the autoreservoir neural network (ARNN) (*35*), the spatiotemporal information conversion machine (STICM) (*36*), and the multitask Gaussian process regression machine (MT-GPRM) (*37*), have been developed for predicting short-term time series within the framework of STI transformation, which depends on an explicit target variable from high-dimensional data.

In this study, according to the STI transformation but relying on a latent variable, we develop a novel ultralow-dimensionality reduction framework: spatial-temporal principal component analysis (stPCA), to fully represent the dynamics of high-dimensional time-series data by only a single latent variable. This framework enables effective detection of the impending critical transitions in dynamical systems with codim-1 local bifurcation by examining the fluctuation of this single variable derived by transforming high-dimensional spatial information into low-dimensional temporal information. On the basis of a solid theoretical background in nonlinear dynamics and delay embedding theory (Fig. 1A), stPCA naturally imposes constraints on samples from the inherent temporal characteristics of dynamical systems, rendering it an applicable method in representing a complex dynamical system without information loss. Furthermore, this single representation variable is analytically obtained from a delay embedding-based optimization problem by solving a characteristic equation, rather than using an iterative numerical optimization algorithm which generally depends on initial values of parameters $W$ (*38*), and can theoretically preserve the dynamical properties of the original high-dimensional system.

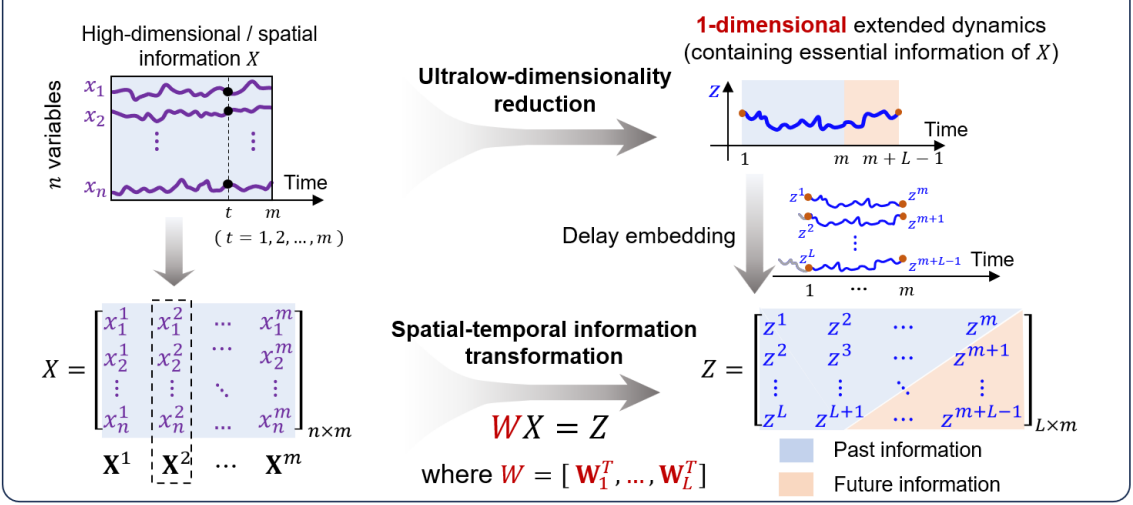
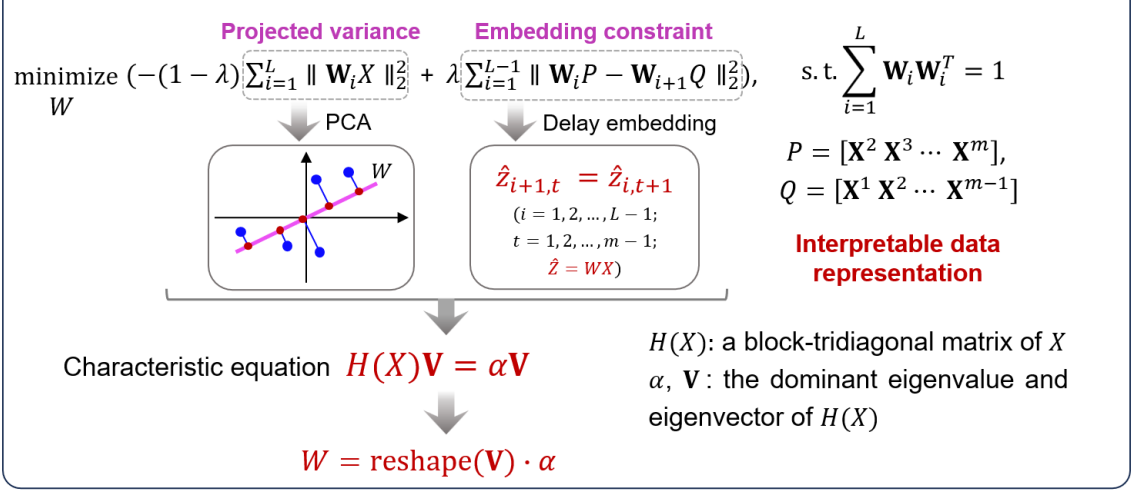
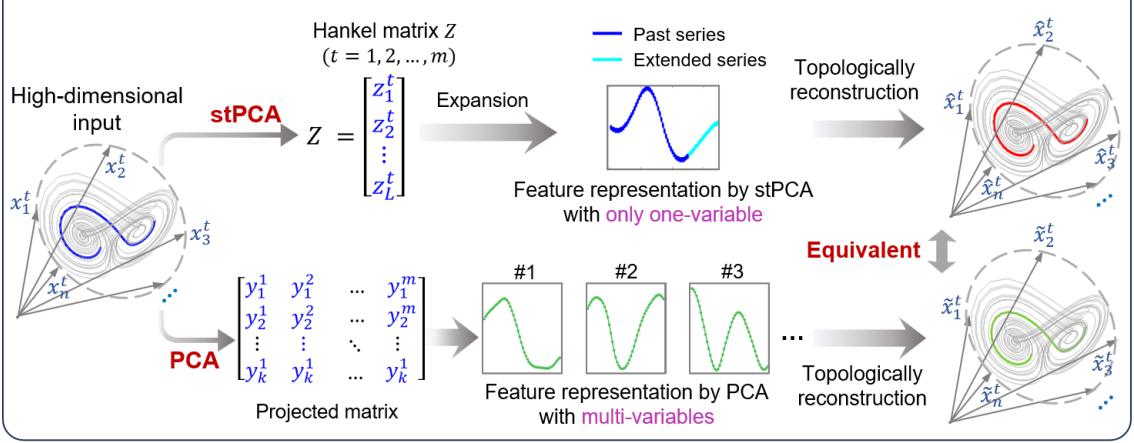

**Figure 1. Schematic illustration of stPCA.** (A) The given high-dimensional "spatial" information $\mathbf{X}^t$ can be transformed into one-dimensional "temporal" information $\mathbf{Z}^t$ based on the STI equations derived from the delay-embedding theorem. (B) Given the high-dimensional time series $X$, stPCA seeks to obtain the transformation matrix $W$ and the projected Hankel matrix $Z$. The optimization objective consists of two terms: the first term aims to maximize the variance of the projected variable $Z$, and the second term ensures that the projected Hankel matrix $Z$ satisfies the delay embedding condition. The analytical solution to the optimization problem is obtained by solving a characteristic equation $H(X)\mathrm{V} = \alpha \mathrm{V}$, where $H(X)$ is a block-tridiagonal matrix. (C) Through stPCA, the high-dimensional time series $X$ is effectively reduced to the Hankel matrix $Z$, which is actually formed by a single variable. On the other hand, PCA is employed to reduce the high-dimensional time series $X$ to $Y$. Notably, both the single latent variable $z$ from stPCA and multiple latent variables from PCA can topologically reconstruct the original dynamical system.

Specifically, stPCA employs a spatial-temporal transformation denoted as $W$, merging (i) the attributes of linear interpretation from PCA and (ii) the temporal dynamics from delay embeddings (Fig. 1B). On one hand, similar to PCA, stPCA requires the maximum projected variance to attain a proper feature transformation (i.e., explainable features), ensuring the capability of a latent variable to quantify critical state transitions based on CSD theory; on the other hand, it requires the projected latent variable to adhere to the delay embedding constraints (i.e., temporal features) (*35, 36*); thus, this ultralow 1-dimensional variable can be theoretically used to topologically reconstruct the original high-dimensional dynamical system. Therefore, it can be obtained by simultaneously optimizing the projected variance and the consistency with the delay embedding for the representative data. The existence of a closed-form solution makes stPCA analytically tractable and offers a fast computational way to even large datasets.

**Results**

**The stPCA framework for high-dimensional time-series data: an analytical and explainable approach of a dynamical system**

The proposed stPCA framework is a dynamic dimensionality reduction method designed to maintain the interpretability of a linear subspace while representing a high-dimensional dynamical system with a single projected variable using a delay embedding strategy. For an observed high-dimensional time series $\mathbf{X}^t = (x_1^t, x_2^t, \ldots, x_n^t)'$ with $n$ variables and $t = 1, 2, \ldots, m$, we construct a corresponding delayed vector $\mathbf{Z}^t = (z^t, z^{t+1}, \ldots, z^{t+L-1})'$ by a delay-embedding strategy with $L > 1$ as the embedding dimension (Fig. 1A), where the symbol " $'$ " represents the transpose of a vector. Clearly, $\mathbf{X}^t$ is a spatial vector with $n$ variables observed at one time point $t$, while $\mathbf{Z}^t$ is a temporal vector formed by only one variable $z$ but at many time points $t, t+1, \ldots, t+L-1$. According to Takens' embedding theory and its generalized versions, such a delay-embedding scheme, $\mathbf{Z}^t$ topologically reconstructs the equivalent dynamics of the original system $\mathbf{X}^t$ if $L > 2d > 0$, where $d$ is the box-counting dimension of the attractor of $\mathbf{X}^t$ (*35-37*), via the spatial-temporal information transformation (STI) equations $\Phi(\mathbf{X}^t) = \mathbf{Z}^t$ shown in the Materials and Methods section. In this way, $Z = (\mathbf{Z}^1, \mathbf{Z}^2, \ldots, \mathbf{Z}^m)$ is a Hankel matrix formed by an extended vector $\mathbf{z} = (z^1, z^2, \ldots, z^m, z^{m+1}, \ldots, z^{m+L-1})$ of a single variable. Then, aiming to identify the critical state transition of a complex dynamical system, we linearize the STI equation as $Z = WX$. Based on the delay embedding requirement for $Z$, the latter $m-1$ elements in the $i$-th row are identical to the former $m-1$ elements in the $(i+1)$-th row of the matrix $Z$, thus we can establish the following equation:

$$\mathbf{W}_i P = \mathbf{W}_{i+1} Q, \tag{1}$$

where $P$ and $Q$ are two submatrices of $X$, i.e., $P = [\mathbf{X}^2 \, \mathbf{X}^3 \, \cdots \, \mathbf{X}^m]$, $Q = [\mathbf{X}^1 \, \mathbf{X}^2 \, \cdots \, \mathbf{X}^{m-1}]$, and $\mathbf{W}_i = (w_{i1}, w_{i2}, \cdots, w_{in})$, $i = 1, 2, \cdots, L-1$ (see the Materials and Methods section for details). More explanations about linear embedding are provided in the Discussion section.

On the other hand, to obtain $W$ such that $Z$ can be obtained through $Z = WX$ given $X = (\mathbf{X}^1 \, \mathbf{X}^2 \, \cdots \, \mathbf{X}^m)$ with $n$ variables $\{x_1^t, x_2^t, \ldots, x_n^t\}$, i.e., $\mathbf{Z}_i = \mathbf{W}_i X$, it is necessary to minimize the reconstruction error or maximize the variance of the projected variable $\|\mathbf{Z}_i\|_2^2$. Then, the sample mean of $X$ is removed as is done in PCA. Thus, the loss function is shown as

$$\underset{W}{\text{minimize}} \quad -(1-\lambda) \sum_{i=1}^{L} \|\mathbf{W}_i X\|_2^2 + \lambda \sum_{i=1}^{L-1} \|\mathbf{W}_i P - \mathbf{W}_{i+1} Q\|_2^2, \tag{2}$$

$$s.t. \sum_{i=1}^{L} \mathbf{W}_i \mathbf{W}_i^T = 1,$$

where $\lambda$ is a predetermined regularization parameter between 0 and 1. The first term of Eq. (2) is the typical PCA loss, which guarantees the maximum projected variance and enables identification the critical state transition, as detailed in Materials and Methods section. The second term makes $Z$ a Hankel matrix, i.e., governed by the delay embedding constraint according to Eq. (1). Theoretically, the obtained single variable $z^t$ can be used to topologically reconstruct the dynamics of the original system $\mathbf{X}^t$, thus characterizing the dynamical features of the whole original system.

Given $X$, in this work, we present an analytical solution to the optimization problem in Eq. (2) by solving the following characteristic equation:

$$H(X)\mathbf{V} = \alpha \mathbf{V}, \tag{3}$$

Page 7

where $H$ is a function that converts $X$ to a block-tridiagonal matrix, $\alpha$ and $\mathbf{V}$ denote the dominant eigenvalue and a corresponding eigenvector of matrix $H(X)$, respectively. Matrix $W$ is obtained by reshaping vector $\mathbf{V}$ (Fig. 1B), then leading to Hankel matrix $Z$ (Fig. 1A). The main time cost comes from solving the dominant eigenvalue of $H(X)$. Utilizing the tridiagonal matrix algorithm (53), its time complexity is $O(nL)$, which ensures the high efficiency of the stPCA algorithm.

**Ultralow-dimensionality data representation and high dynamic robustness: application of stPCA to time-series data of a coupled Lorenz model**

To illustrate the mechanism of the proposed framework, a series of coupled Lorenz models (46, 47)

$$\dot{\mathbf{X}}(t) = G(\mathbf{X}(t); \mathbf{P}) \tag{4}$$

were employed to generate synthetic time-series datasets under different noise conditions, where $G(\cdot)$ is the nonlinear function of the Lorenz system with $\mathbf{X}(t) = (x_1^t, \ldots, x_n^t)'$, $\mathbf{P}$ is a parameter vector, and $n$ is the dimension of $\mathbf{X}(t)$. More details about the Lorenz system are provided in SI Appendix Note S4.

To study the intrinsic dynamical information of variable $z$ via stPCA, single value decomposition (SVD) is carried out on the Hankel matrix $Z$ stacked by $z$:

$$Z = \begin{pmatrix} z^1 & z^2 & \cdots & z^m \\ z^2 & z^3 & \cdots & z^{m+1} \\ \vdots & \vdots & \ddots & \vdots \\ z^L & z^{L+1} & \cdots & z^{m+L-1} \end{pmatrix}_{L \times m} = U\Sigma R. \tag{5}$$

The columns of matrices $U$ and $R$ after SVD are structured hierarchically according to their capability to represent the columns and rows of $Z$. Usually, $Z$ can be derived via a low-rank approximation by utilizing the first $r$ columns from $U$ and $R$ (20, 21). The initial $r$ columns of $R$ represent a time series of the intensity of each column of $U\Sigma$ (56). In this manner, we can decompose the single variable $z$ into multiple principal components projections (Fig. 2C, 2D).

To quantify the similarity between noisy cases and noise-free cases for the principal component projections of the stPCA and PCA algorithms respectively, we designed a metric called the principal component Frechet distance (PCFD) based on the Frechet distance (54, 55) (see SI Appendix Note S7 for details). Given a metric space $(\mathcal{H}, D)$ with $D$ as the distance metric in $\mathcal{H}$, for two curves $y: [b_1, b_2] \to \mathcal{H}$ and $z: [b_1', b_2'] \to \mathcal{H}$, their discrete Fréchet distance (DFD) is given as

$$\text{DFD}(y, z) = \inf_{\beta_y, \beta_z} \max_{t \in [0,1]} D(y(\beta_y(t)), z(\beta_z(t))), \tag{6}$$

where $\beta_y$ and $\beta_z$ are arbitrary continuous nondecreasing functions from $[0,1]$ onto $[b_1, b_2]$ and $[b_1', b_2']$, respectively. To better evaluate the distance of the two projection trajectories, we normalize the curves/projections by $\tilde{y} = \text{Normalize}(y)$ and $\tilde{z} = \text{Normalize}(z)$, then

$$\text{PCFD}(y, z) = \min\{\text{DFD}(\tilde{y}, \tilde{z}), \text{DFD}(-\tilde{y}, \tilde{z})\}. \tag{7}$$

By plotting the projections of the first three columns of matrix $R$ in Eq. (5) of one noise-free case ($m = 50, L = 20, n = 20, \sigma = 0$), we found that the projections of the top principal components from $Z$ by stPCA are very similar to those by PCA (SI Appendix Fig. S5), with PCFD values of 0.108, 0.291 and 0.254 for the top 3 components, respectively, which indicates that the 1-dimensional variable $z$ actually contains as much information as multiple projected variables from PCA. This characteristic is evidence that our stPCA method achieves efficient dimensionality reduction to an ultralow-dimensional space without information loss compared with traditional PCA. The analysis results obtained by stPCA and PCA for more cases are presented in SI Appendix Fig. S1. Furthermore, both stPCA and PCA were applied in two scenarios: one with the original sample order and another with a randomly shuffled sample order, as shown in SI Appendix Fig. S2.

In addition, we discuss the situation when there is additive noise, where $n = 60, m = 50, L = 20$, and noise intensity $\sigma = 20$ (Fig. 2A). The results are illustrated in Figs. 2B-2I. The projections of the top three principal components from stPCA in the presence of noise closely resemble those from the noise-free situation which are shown in SI Appendix Fig. S5. Besides, Fig. 2D and Fig. 2H demonstrate the high robustness of stPCA against noise in contrast to PCA, which may be due to the inherent temporal constraints of stPCA imposed by the dynamical system in a steady state. The results of more cases with different noise intensities are illustrated in SI Appendix Fig. S5. The quantified PCFD values for different embedding dimensions $L$ are presented in SI Appendix Table S1.

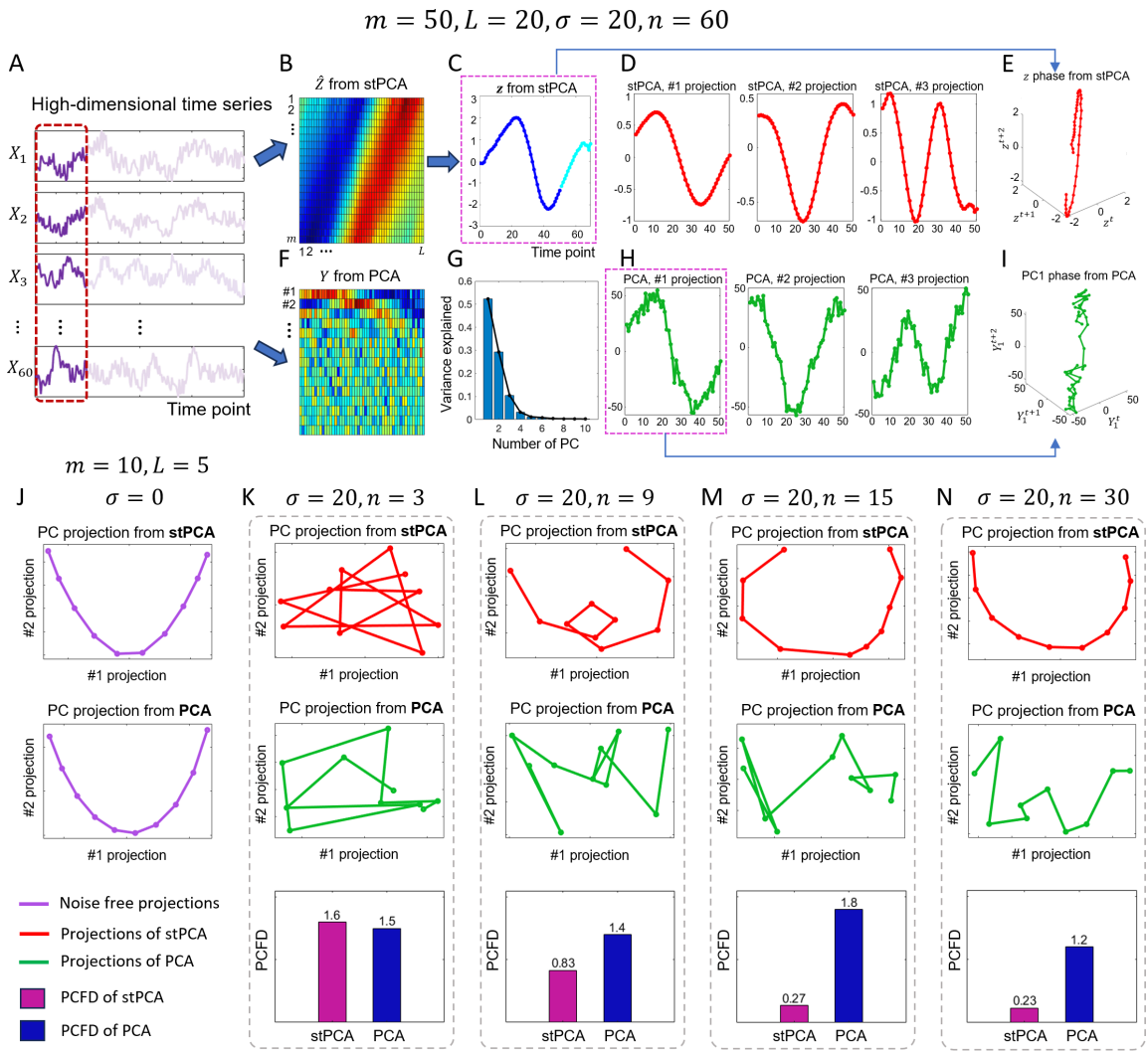

**Figure 2. Synthetic time-series datasets of a coupled Lorenz model were generated in noise-free and noisy situations.** (A)-(I) For the strong noise-perturbed scenario (noise intensity $\sigma = 20$), the dimensionality reduction results of a time series with length $m = 50$ and embedding dimension $L = 20$ of the 60D-coupled ($n = 60$) Lorenz system. (A) The original high-dimensional time series $X$. (B) The heatmap of matrix $\hat{Z}$ obtained by stPCA. (C) The one-dimensional variable $z$ obtained by stPCA. (D) Projections of top three principal components (denoted as PC for short in the figure) obtained from the Hankel matrix from stPCA. (E) The projected phase of the one-dimensional variable $z$. (F) The heatmap of matrix $Y$ obtained by PCA. (G) The proportions of PC variances from PCA. (H) Projections of top three PCs from PCA. (I) The projected phase of the #1 PC projection from PCA. (J) The projections of top 2 PCs from stPCA or PCA for a low-dimensional system ($n = 3$) with $m = 10, L = 5$ and noise intensity $\sigma = 0$. (K)-(N) For dimensionalities $n = 3, 9, 15$ and $30$, the curves of top 2 PC projections from stPCA and PCA, along with the PCFD between the curve in Fig. 2J and that in the strong noise-perturbed scenario by stPCA and PCA, respectively.

We also applied stPCA and PCA to short-term time series with $m = 10$ and $L = 5$. With the relatively low dimensionality $n = 3$ and a noise intensity $\sigma = 0$ of system Eq. (4), the projections of the top two principal components obtained from stPCA and PCA are illustrated in Figs. 2J. When strong noise is introduced ($\sigma = 20$), we illustrated the top two principal components for dimensionalities $n = 3, 9, 15$ and $30$ in Figs. 2K-2N. Clearly, with low dimensionality ($n = 3$), both stPCA and PCA exhibit unsatisfactory performance due to the presence of noise compared with the noise-free situation (Fig. 2K). As dimensionality increases, the #1-#2-projection curve obtained from stPCA becomes smoother, and the PCFD between this curve and that of stPCA in Fig. 2J decreases from 1.6 to 0.23 (Figss. 2L-2N). Meanwhile, the projection curve from PCA always exhibits irregular fluctuations owing to noise perturbations. The improved performance of stPCA with the increasement of dimensionality of system Eq. (4) indicates that the inclusion of additional relevant variables results in an expansion of spatial information. According to Takens' embedding theory, such extra/increased spatial information can be transformed into temporal information, thereby enhancing the robustness of stPCA to noise, which also demonstrates the ability of stPCA in converting spatiotemporal information. More results are demonstrated in SI Appendix Fig. S3. We also applied stPCA to longer time series of a coupled Lorenz system and the DREAM4 dataset, and the results are shown in Figs. S4 and S7.

**The early-warning signals of a critical transition: application of stPCA to the tipping point detection of multiple-node networks**

To illustrate how stPCA effectively characterizes high-dimensional time series, two network models were employed (*47*): a sixteen-node/variable system with a Fold bifurcation (Fig. 3A) and an eighteen-node/variable system with a Hopf bifurcation (Fig. 3E). More details of these two networks are presented in SI Appendix Figs. S8, S9 and in Note S5. From these systems, two datasets were generated by varying parameter $\tau$ from 1 to 310 and from 1 to 190, respectively (SI Appendix Tables S2 and S3). The critical transition points were set at $\tau = 210$ (Hopf bifurcation) and $\tau = 100$ (fold bifurcation). To investigate different states with the varying parameters of a dynamical system, a sliding-window scheme was implemented to divide the entire process into smaller windows. For each window, the dynamics of the high-dimensional time series are reduced to that of a latent variable $z$ from stPCA (Figs. 3B and 3F). Based on the CSD or dynamic network biomarker (DNB) (*5*), the abrupt increase in the standard deviation (SD) of $\mathbf{z} = (z^1, z^2, \cdots, z^m, z^{m+1}, \cdots, z^{m+L-1})$ (or the fluctuations of the '$z$' dynamics in a sliding window) indicates the upcoming critical transition (Figs. 3C and 3G). The details are shown in the Materials and Methods section.

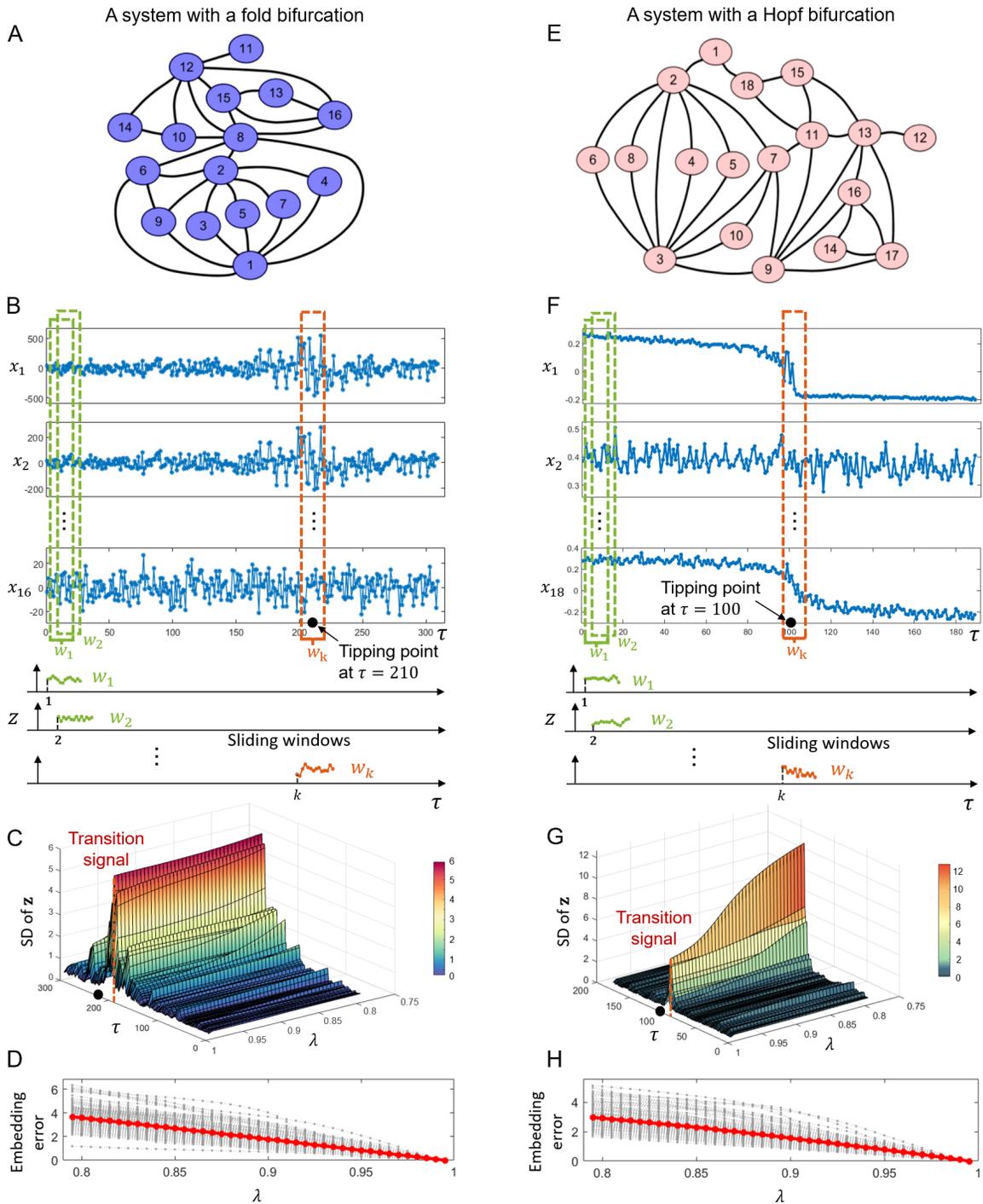

**Figure 3. Application of stPCA to the tipping point detection in simulated datasets.** The datasets are generated from a multiple-node network model. **(A)** A 16-node model of a 16-dimensional dynamical system with a Hopf bifurcation. **(B)** The observed dynamics of each node/variable as parameter $\tau$ varies for the 16-node model. To capture the early-warning signals of the tipping points before the critical transitions of the dynamical system, the original high-dimensional time series are partitioned into sliding windows. The one-dimensional latent variable $z$

is obtained by stPCA from each sliding window. **(C)** In each sliding window, the standard deviation (SD) of the latent variable $z$ is calculated with the varying parameter $\tau$ and regularization parameter $\lambda$ for the 16-node model. **(D)** The red and gray curves represent the average embedding error and all embedding errors as $\lambda$ varies, respectively. **(E)** An 18-node model of an 18-dimensional dynamical system with a fold bifurcation. **(F)** The observed dynamics of each node/variable as parameter $\tau$ varies for the 18-node model. The one-dimensional latent variable $z$ is obtained by stPCA from each sliding window. **(G)** The SD of $z$ is calculated within a sliding window as parameters $\tau$ and $\lambda$ varies for the 18-node model. **(H)** The red curve represents the curve of the average embedding error as $\lambda$ varies, while the gray curves represent the curve of all embedding errors as $\lambda$ varies for the 18-node model.

In addition, when the regularization parameter $\lambda$ varies from 0.78 to 0.98, stPCA can effectively detect the imminent critical transitions of systems with fold bifurcation (Fig. 3C) and Hopf bifurcation (Fig. 3G) through the one-dimensional dynamics of $z$. However, the embedding error $E_{embedding} = \parallel Z - \hat{Z} \parallel_F$ increases accordingly (Figs. 3D and 3H) with decreasing $\lambda$. To meet the delay embedding condition so that $\hat{Z}$ can reconstruct the original high-dimensional dynamics, it is necessary to keep the embedding error below 2, which suggests that $\lambda$ typically ranges from 0.9 to 1.

**ICU decision-making: real-world application of stPCA on MIMIC datasets**

The MIMIC-III (Medical Information Mart for Intensive Care III) database (*50*) is a widely used resource for studying critical care patients (*41-45*). It comprises deidentified health-related data associated with patients who stayed in the critical care units of the Beth Israel Deaconess Medical Center between 2001 and 2012 and contains comprehensive clinical information collected from the ICU stays (*60, 62*). The percentage of patients who experienced death within a 5-year period was 2.3%, and unplanned 30-day readmissions were 12.9%; therefore, making accurate judgments about whether a patient is ready for discharge from the ICU is of utmost importance in optimizing life-saving measures with the constraints of available medical resources. By analyzing these heterogeneous high-dimensional noisy data, we aimed to gain insights into the complex nature of critical care conditions and contribute to improving patient management and decision-making processes.

In this study, we first applied stPCA to a subset of the MIMIC-III database. We randomly selected 10,000 patients from this database. After a rigorous preprocessing procedure, which involved data cleaning and quality control measures, a final dataset comprising 2,908 patients was obtained (Fig. 4A). The data of each patient were represented by a heterogeneous high-dimensional time series, i.e., a matrix consisting of at least 10 time points indicating the number of hours since his/her admission and containing no less than five diagnosis-related indicators. Additional details of the preprocessing process are provided in SI Appendix Fig. S11.

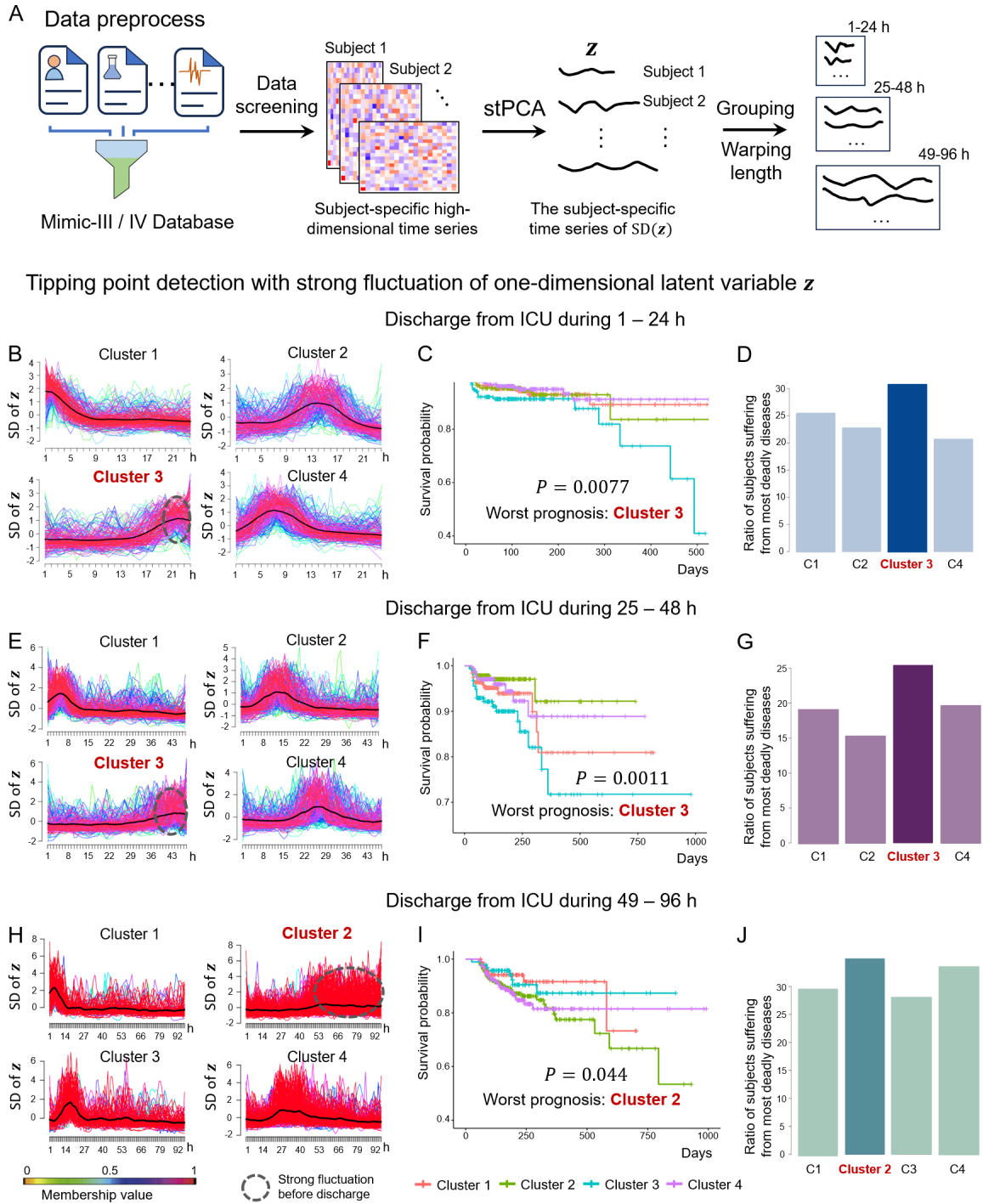

**Figure 4. Performance of stPCA applied on the MIMIC-III dataset with three groups on patients' ICU stay. (A)** Patient admission data were extracted from

Page 14

*Science Advances*

the ICU PDMS. According to the screening criteria, we obtained 2908 patient-specific matrices that represent tested disease indicators at different hours. The patients were divided into three groups, with **(B)(C)(D)** representing patients who stayed in the ICU for 0-24 hours, **(E)(F)(G)** representing patients who stayed in the ICU for 25-48 hours, and **(H)(I)(J)** representing patients who stayed in the ICU for 49-96 hours. For each group, high-dimensional disease indicators of each patient were dimensionally reduced to a single variable by stPCA. Then, a DTW scheme was adopted to align these time series so that they have the same length. **(B)** Each colored curve corresponds to the SD of the one-dimensional variable for each patient, and the black curve represents the average SD. Then, time series clustering was performed on these curves, resulting in four clusters. Notably, Cluster 3 exhibits strong fluctuations in the later stages. **(C)** Survival analysis results for the four clusters with an ICU duration of 0-24 hours. **(D)** Proportion of patients in the four clusters with an ICU duration of 0-24 hours who experienced the highest disease mortality rate. **(E)** Time series clustering results for patients with an ICU duration of 25-48 hours, with Cluster 3 exhibiting strong fluctuations in the later stages. **(F)** Survival analysis results for the four clusters with ICU durations of 25-48 hours. **(G)** Proportion of patients in the four clusters with ICU durations of 25-48 hours who experienced diseases with the highest mortality rates. (**H**) Time series clustering results for patients with ICU durations of 49-96 hours, with Cluster 2 exhibiting strong fluctuations in the later stages. **(I)** Survival analysis results for the four clusters with ICU durations of 49-96 hours. **(J)** Proportion of patients in the four clusters with ICU durations of 49-96 hours who experienced diseases with the highest mortality rates.

For each patient, the high-dimensional time series was reduced to one variable $z$ by stPCA with the sliding-window scheme, and then the fluctuation of $z$, $Fl^z$, was calculated; i.e., $Fl^z_{(k)} = \text{SD}(z^1_{(k)}, z^2_{(k)}, \cdots, z^m_{(k)}, z^{m+1}_{(k)}, \cdots, z^{m+L-1}_{(k)})$ was calculated for the $k$th sliding window (see SI Appendix Fig. S12). Based on the individual variations in ICU stay duration, we classified the curves of $Fl^z$ into four groups: Group 1 (ICU stay duration 1-24 hours), Group 2 (ICU stay duration 25-48 hours), Group 3 (ICU stay duration 49-96 hours), and Group 4 (ICU stay duration exceeding 4 days). To explore underlying dynamic patterns among patients in each group, we first used the dynamic time warping (DTW) algorithm (*63*, *64*) to align the curves of $Fl^z$ so that they have a consistent length. The details of the DTW algorithm are presented in SI Appendix Note S8.

Then, the SD curves were clustered (*64*, *65*) within each group based on the aligned time series. Four clusters were obtained within each group (Figs. 4B, 4E and 4H). Notably, there is a common pattern among Cluster 3 of Group 1 (Fig. 4C), Cluster 3 of Group 2 (Fig. 4F), and Cluster 2 of Group 3 (Fig. 4I), all of which exhibit relatively strong fluctuations in $z$ in the later stages, correlating with the poorest recorded prognosis within the corresponding cluster. In addition, there was a significant difference ($\text{pvalue} < 0.05$) in the prognoses of patients among different clusters in each group. Furthermore, by examining the proportion of patients within each cluster who suffered from diseases/diagnoses with the highest mortality rates (Figs. 4D, 4G and 4J), it was found that in these clusters with poor prognoses, the proportions of patients afflicted by diagnoses with high mortality rates were also higher. Based on these observations, we speculate that for an individual patient, the significant fluctuations of $z$ from stPCA may indicate a critical condition, warranting close monitoring and ongoing treatment until the $Fl^z$ index stabilizes.

Furthermore, on the basis of $Fl^z$, i.e., the fluctuation of the latent variable $z$, and 2~5 diagnosis-specific indicators, we proposed a new scheme to determine whether a patient should be discharged from the ICU at time point $t$, i.e., ICU decision-making, as follows:

In recent hours, $\{t - wl + 1, \ldots, t - 1, t\}$, where $wl$ represents the window length, if the average $Fl^z$ for a patient is significantly lower than the highest value observed since admission, it indicates a relatively stable condition after treatment. An index $\text{idx}_z(t)$ of $z$ is defined as follows:

$$\text{idx}_z(t) = \max(\text{mean}(\mathbf{S}_{\text{past}(j)})) / \text{mean}(\mathbf{S}_{\text{recent}}), \tag{8}$$

where vector $\mathbf{S}_{\text{past}(j)}$ represents a vector of standard deviations of $z$ during a past period (time window $j$); i.e., $\mathbf{S}_{\text{past}(j)} = [Fl^z_{(j-wl+1)}, \ldots, Fl^z_{(j-1)}, Fl^z_{(j)}]$, with $j = wl, wl + 1, \ldots, t - 1$. In addition, vector $\mathbf{S}_{\text{recent}}$ represents the standard deviations of $z$ over the most recent $wl$ time points, i.e., $\mathbf{S}_{\text{recent}} = [Fl^z_{(t-wl+1)}, \ldots, Fl^z_{(t-1)}, Fl^z_{(t)}]$.

During the recent period, 2-5 diagnosis-specific items $x_i^t$ that are most closely associated should fall within the normal range. This can be evaluated using the following condition: $\text{itmFlg}(t) = \sum_{i=1}^{K} \#(x_i^t)$ with item flags

$$\#(x_i^t) = \begin{cases} 1, & Lb_i \leq x_i^t \leq Ub_i, \\ 0, & \text{otherwise,} \end{cases}$$

where $K$ is the number of selected items, $Lb_i$ and $Ub_i$ represent the lower and upper bounds of the normal range for $x_i^t$, respectively.

Then, the decision for ICU discharge at time point $t$ is described as follows:

$$\text{Decision}(t) = \begin{cases} 1, & \text{idx}_z(t) \geq FC \text{ and } \text{itmFlg}(t) = K, \\ 0, & \text{otherwise,} \end{cases} \tag{9}$$

where $FC$ represents the fold change threshold. If $\text{Decision}(t) = 1$, i.e., a positive decision at time $t$, then the patient's condition is stable at time point $t$, thereby indicating suitability for ICU discharge. Otherwise, if $\text{Decision}(t) = 0$, i.e., a negative decision at time $t$, then the patient should stay in the ICU for further treatment and monitoring. An illustration of this discharge scheme is presented in SI Appendix Fig. S13.

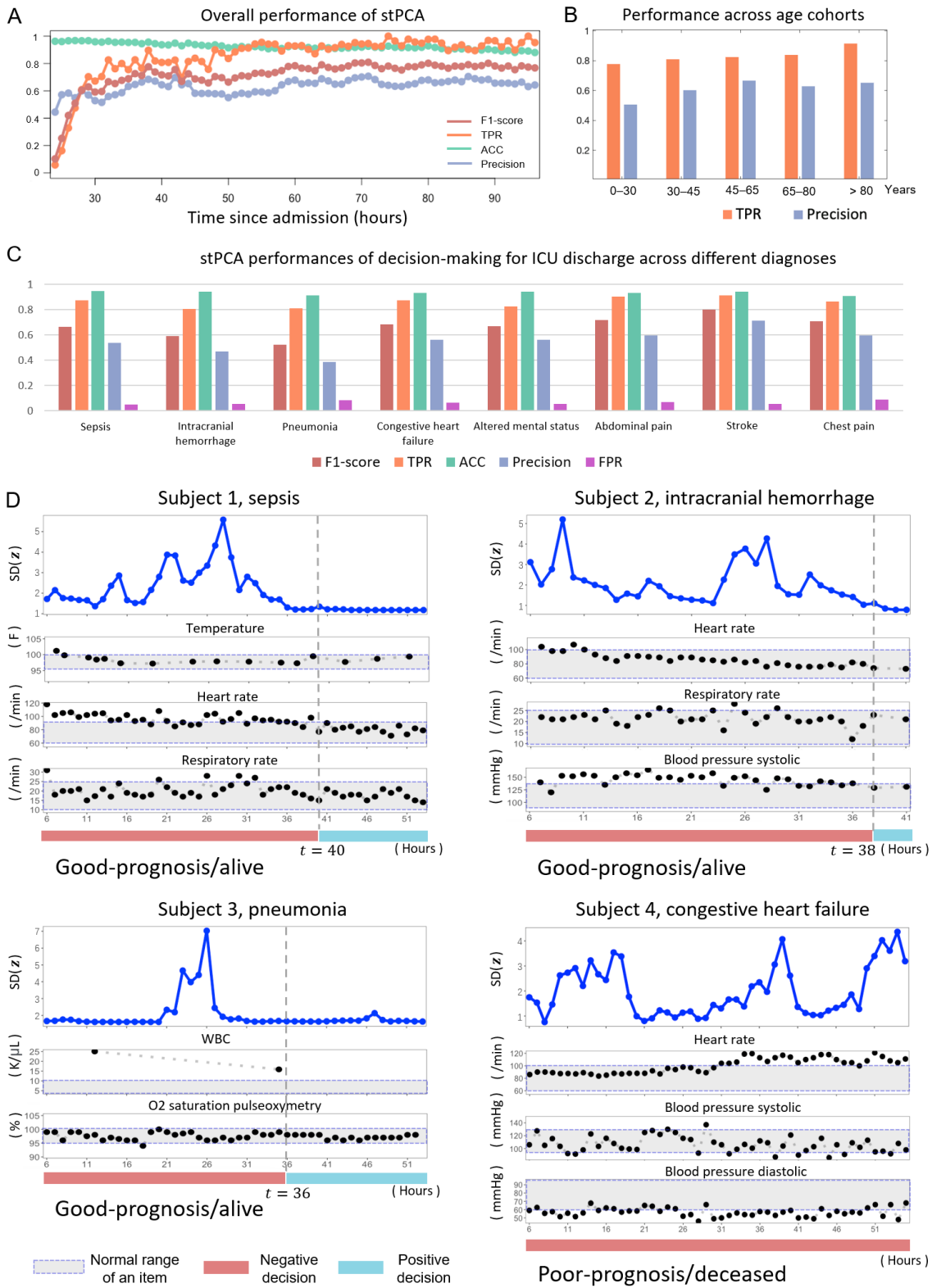

**Figure 5. Performance of the decision for ICU discharge.** (**A**) For all samples, the performance of stPCA-based discharge decision was assessed for each hour

since admission. (**B**) TPR/Recall and Precision across five age cohorts are presented. (**C**) For the top eight diagnoses with the highest mortality rate, we evaluate the performance of our discharge decision metrics based on the stPCA algorithm. (**D**) We present four typical cases (patients suffering from sepsis, intracranial hemorrhage, pneumonia, and congestive heart failure). Combining the dimensionality reduction results of stPCA and 2-5 diagnosis-related indicators, we can make decisions on whether patients should be discharged from the ICU. The red interval indicates that the patient should continue receiving ICU treatment or observation, while the blue interval suggests that the patient's condition is relatively stable and that they can be discharged from the ICU.

A set of standardized metrics are employed for evaluating the performance of the ICU discharge decision for a specific method as follows. When $\text{Decision}(t) = 1$, if the patient is discharged during a period of 5 hours after $t$ or all the relevant items thereafter are within the normal range, then this decision is considered a true positive (TP); otherwise, it is a false-positive (FP). When $\text{Decision}(t) = 0$, if the patient is currently receiving treatment in the ICU, then this decision is considered a true negative (TN); otherwise, this decision is considered a false negative (FN). The F1-score, accuracy (ACC), true positive rate (TPR)/recall, false positive rate (FPR), and precision metrics are calculated based on TP, FP, TN, and FN. The detailed descriptions and calculation formulas for these metrics are provided in SI Appendix Note S6.

Based on the aforementioned scheme, the overall performance of stPCA-based decision-making was assessed over the course of time since admission (Fig. 5A). The precision and F1-score are around 0.6 and 0.7, while TPR/recall, and ACC metrics are all greater than 0.8 when $t \geq 30\,\text{h}$. When $t < 30\,\text{h}$, the slightly lower F1-score and TPR are due to the limited number of discharged patients. Then, the performances across five age cohorts are demonstrated in Fig. 5B. Moreover, Fig. 5C presents the decision performance of stPCA for the top-8 diagnoses with the highest mortality rates: sepsis, intracranial hemorrhage, pneumonia, congestive heart failure, altered mental status, abdominal pain, stroke, and chest pain. Across various metrics, our decision scheme based on stPCA consistently demonstrates favorable results. Figure 5D illustrates the assessment of ICU discharge for four subjects diagnosed with sepsis, intracranial hemorrhage, pneumonia, and congestive heart failure. When $t \geq 40$ h, 38 h, and 36 h for Subjects 1, 2, and 3, respectively, we determined that $\text{Decision}(t) = 1$; thus, Subjects 1-3 met the criteria for ICU discharge after treatment, which is consistent with their positive prognoses, as recorded. Conversely, Subject 4 exhibited inadequate treatment response and experienced a deteriorating state over time, which is consistent with the recorded negative prognosis.

## Performance comparisons of dimensionality reduction methods on the MIMIC-III and MIMIC-IV datasets

To further evaluate the performance of the stPCA algorithm in ICU decision-making, stPCA is compared with seven commonly used dimensionality reduction or data representation methods including (i) principal component analysis (PCA) (*11*), (ii) locally linear embedding (LLE) (*9*), (iii) t-distributed stochastic neighbor embedding (t-SNE) (*6*), (iv) multidimensional scaling (MDS) (*8*), (v) isometric feature mapping (ISOMAP) (*10*), (vi) autoencoder (AE) (*19*), (vii) variational autoencoder (VAE) (*66*), MARBLE(*25*), CANDyMan(*24*), HAVOK(*28*), which can potentially be applied to the decision-making process of ICU patient discharge. The seven methods for comparison are listed below, and the details are presented in SI Appendix Note S9. On the MIMIC-III data, the results for all samples and the top-8 high-mortality diagnoses are illustrated in Table 1. Clearly, stPCA consistently achieves the best performance among all methods. The comparisons of time complexity between deep learning-based representation learning methods and stPCA are provided in Supplementary Information.

**Table 1.** Comparison of discharge-decision performances on the MIMIC-III dataset for stPCA and 10 other dimensionality reduction methods

| Method | Diagnosis / Metrics | All samples | Sepsis | Intracranial hemorrhage | Pneumonia | Congestive heart failure | Altered mental status | Abdominal pain | Stroke | Chest pain |
|---|---|---|---|---|---|---|---|---|---|---|
| stPCA | Recall | **0.835** | 0.869 | 0.803 | 0.810 | 0.869 | 0.825 | 0.902 | 0.911 | 0.863 |
|  | Precision | **0.639** | 0.534 | 0.467 | 0.386 | 0.560 | 0.559 | 0.593 | 0.713 | 0.595 |
|  | F1-score | **0.724** | 0.662 | 0.591 | 0.523 | 0.681 | 0.666 | 0.716 | 0.800 | 0.704 |
| PCA | Recall | **0.625** | 0.798 | 0.538 | 0.486 | 0.834 | 0.569 | 0.211 | 0.766 | 0.604 |
|  | Precision | **0.507** | 0.685 | 0.318 | 0.379 | 0.569 | 0.541 | 0.154 | 0.837 | 0.733 |
|  | F1-score | **0.560** | 0.737 | 0.400 | 0.426 | 0.676 | 0.555 | 0.178 | 0.800 | 0.662 |
| LLE | Recall | **0.496** | 0.000 | 0.308 | 0.333 | 0.500 | 0.547 | 0.063 | 0.000 | 0.178 |
|  | Precision | **0.497** | 0.000 | 0.500 | 0.253 | 0.914 | 0.744 | 0.067 | NaN | 0.099 |
|  | F1-score | **0.496** | NaN | 0.381 | 0.288 | 0.646 | 0.630 | 0.065 | NaN | 0.127 |
| t-SNE | Recall | **0.556** | 0.429 | 0.750 | 0.760 | 0.538 | 0.549 | 0.714 | 0.519 | 0.593 |
|  | Precision | **0.476** | 0.178 | 0.468 | 0.442 | 0.574 | 0.196 | 0.507 | 0.400 | 1.000 |
|  | F1-score | **0.513** | 0.252 | 0.576 | 0.559 | 0.555 | 0.289 | 0.593 | 0.452 | 0.745 |
| MDS | Recall | **0.196** | 0.000 | 0.000 | 0.000 | 0.267 | 0.000 | 0.000 | 0.000 | 0.116 |
|  | Precision | **0.471** | 0.000 | NaN | 0.000 | 1.000 | NaN | NaN | NaN | 0.385 |
|  | F1-score | **0.277** | NaN | NaN | NaN | 0.421 | NaN | NaN | NaN | 0.178 |
| ISOMAP | Recall | **0.257** | 0.257 | 0.182 | 0.494 | 0.396 | 0.468 | 0.000 | 0.400 | 0.269 |
|  | Precision | **0.495** | 0.205 | 0.500 | 0.534 | 0.875 | 0.400 | 0.000 | 0.615 | 0.359 |
|  | F1-score | **0.338** | 0.228 | 0.267 | 0.513 | 0.545 | 0.431 | NaN | 0.485 | 0.308 |
| AE | Recall | **0.486** | 0.667 | 0.441 | 0.587 | 0.568 | 0.438 | 0.375 | 0.820 | 0.357 |
|  | Precision | **0.465** | 0.464 | 0.312 | 0.514 | 0.618 | 0.286 | 0.231 | 0.696 | 0.357 |
|  | F1-score | **0.475** | 0.547 | 0.365 | 0.548 | 0.592 | 0.346 | 0.286 | 0.753 | 0.357 |
| VAE | Recall | **0.670** | 0.767 | 0.561 | 0.598 | 0.819 | 0.554 | 0.516 | 0.772 | 0.576 |
|  | Precision | **0.587** | 0.439 | 0.535 | 0.359 | 0.659 | 0.240 | 0.356 | 0.379 | 0.690 |
|  | F1-score | **0.626** | 0.558 | 0.548 | 0.449 | 0.730 | 0.335 | 0.421 | 0.508 | 0.628 |

\* NaN means there is no such event.

Additionally, we extended the application of our algorithm to an independent database called MIMIC-IV, which comprises medical records corresponding to patients admitted to an ICU or the emergency department between 2008 and 2019 (*51*). Similarly, a cohort of 10,000 patients was randomly selected from this extensive dataset. After subjecting the data to a rigorous preprocessing procedure similar to that for MIMIC-III (Fig. 4A), we curated a focused sub-dataset consisting of 625 patients, based on which the ICU decision-making approach from stPCA was compared with those from seven other data representation methods, which is in line with the above evaluation protocol. The performance metrics for all subjects, as well as those pertaining to the top-5 most prevalent diagnoses, are shown in Table 2. In a general overview, it is noted that our discharge strategy from stPCA consistently outperforms those from other methods. The observation from applications on both the independent MIMIC-III and MIMIC-IV datasets highlights the robustness and effectiveness of the stPCA approach.

**Table 2.** Comparison of discharge-decision performances on the MIMIC-IV dataset for stPCA and 7 other dimensionality reduction methods

| Method | Diagnosis / Metrics | All samples | Acute posthemorrhagic anemia | Pure hypercholesterolemia | Acidosis | Thrombocytopenia | Cardiogenic shock |
|---|---|---|---|---|---|---|---|
| stPCA | Recall | **0.886** | 0.913 | 0.966 | 0.973 | 0.933 | 0.988 |
|  | Precision | **0.676** | 0.668 | 0.824 | 0.636 | 0.617 | 0.837 |
|  | F1-score | **0.767** | 0.772 | 0.889 | 0.769 | 0.743 | 0.906 |
| PCA | Recall | **0.553** | 0.491 | 0.9 | 0.804 | 0.369 | 0.952 |
|  | Precision | **0.684** | 0.358 | 0.006 | 0.643 | 0.346 | 0.900 |
|  | F1-score | **0.611** | 0.414 | 0.885 | 0.714 | 0.358 | 0.925 |
| LLE | Recall | **0.091** | 0.090 | 0 | 0.350 | 0.288 | 0.667 |
|  | Precision | **0.493** | 0.818 | NaN | 0.667 | 0.826 | 0.780 |
|  | F1-score | **0.154** | 0.167 | NaN | 0.459 | 0.427 | 0.719 |
| t-SNE | Recall | **0.671** | 0.743 | 0.923 | 0.76 | 0.586 | 0.927 |
|  | Precision | **0.714** | 0.699 | 0.774 | 0.31 | 0.288 | 0.692 |
|  | F1-score | **0.692** | 0.720 | 0.842 | 0.441 | 0.386 | 0.793 |
| MDS | Recall | **0.317** | 0.340 | 0 | 0.631 | 0.229 | 0.842 |
|  | Precision | **0.631** | 0.776 | 0 | 0.477 | 0.438 | 1 |
|  | F1-score | **0.422** | 0.474 | NaN | 0.543 | 0.301 | 0.914 |
| ISOMAP | Recall | **0.766** | 0.818 | 0.833 | 0.875 | 0.818 | 0.907 |
|  | Precision | **0.783** | 0.886 | 1 | 0.509 | 0.706 | 0.834 |
|  | F1-score | **0.774** | 0.851 | 0.909 | 0.644 | 0.758 | 0.869 |
| AE | Recall | **0.284** | 0.446 | 0.651 | 0.627 | 0.254 | 0.846 |
|  | Precision | **0.709** | 0.631 | 0.757 | 0.646 | 0.340 | 0.733 |
|  | F1-score | **0.405** | 0.522 | 0.700 | 0.636 | 0.291 | 0.786 |
| VAE | Recall | **0.686** | 0.554 | 0.855 | 0.785 | 0.365 | 0.851 |
|  | Precision | **0.634** | 0.663 | 0.776 | 0.646 | 0.265 | 0.776 |
|  | F1-score | **0.659** | 0.603 | 0.813 | 0.709 | 0.307 | 0.814 |

* NaN means there is no such event.

## Identifying the tipping point for single-cell embryonic development

To illustrate how stPCA detects the signal of cell differentiation, the proposed method was applied to an scRNA-seq dataset of embryonic development (*67*) (Fig. 6A). We selected the top 200 highly variable genes, and then reordered 758 single cells along the pseudotime (*68*). Subsequently, the SD of *z* was calculated by stPCA with the sliding window scheme. From Fig. 6B, it is seen that there is an abrupt increase in the SD curve at 36 hours, which signals the cell fate transition of differentiation into definitive endoderm (DE). The identified critical transition is validated by the experimental observation, that is, such signal is notably earlier than the typical emergence of DE cells around days 4 or 5 in established human pluripotent stem cell protocols (*67*).

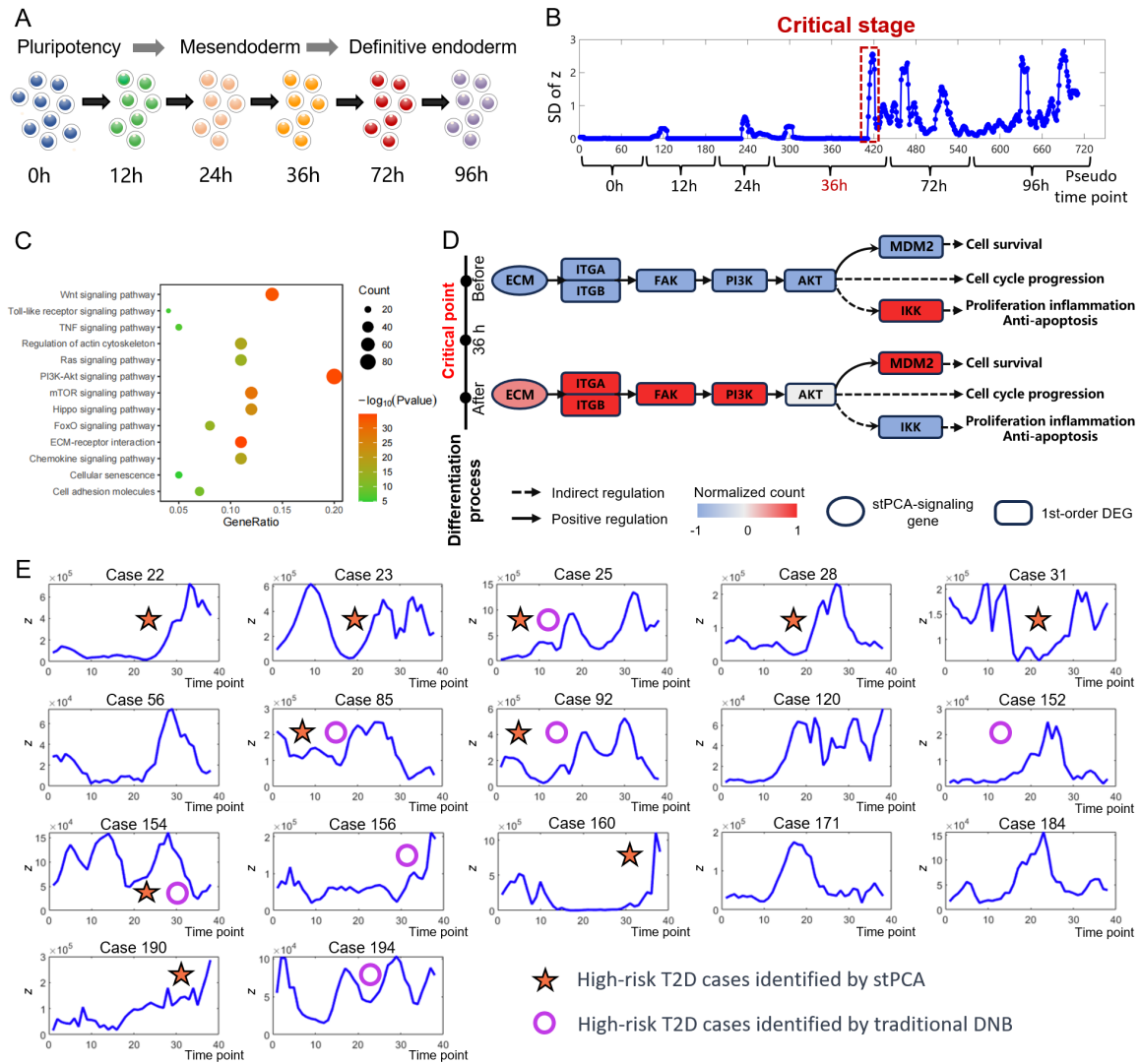

**Figure 6. Detecting the tipping points during embryonic development and analyzing T2D risk by stPCA.** (A) The biological process of cell differentiation from a pluripotent state to mesendoderm and, finally, to definitive endoderm (DE)

cells. (B) The SD curve of the latent variable *z* with the sliding-window scheme. It is shown that there was a sudden increase at 36 hours during the differentiation. (C) The pathway enrichment analysis for the stPCA-signaling genes/DNB. (D) The underlying molecular mechanism revealed by the functional analysis of stPCA-signaling genes and their first-order neighboring differentially expressed genes (DEGs). (E) The SD of *z* for 17 cases confirmed T2D subsequently. The orange star presents a case with a high risk of developing T2D by stPCA, and the blue circle denotes a case with a high risk of developing T2D by traditional DNB.

To investigate the molecular regulation mechanism underlying cell differentiation during embryonic development, the biological functional analyses such as signaling pathway enrichment analysis and gene annotation were performed to reveal the potential biological functions of 50 stPCA-signaling/DNB genes found at 36 hours and their first-order differentially expressed genes (DEGs). The enrichment results showed that these genes were mainly enriched in the Wnt signaling pathway, PI3K-Akt signaling pathway, mTOR signaling pathway, and other differentiation-related signaling pathways (Fig. 6C). The results also show that the synergy of the stPCA-signaling genes and their 1st-order neighboring DEGs regulate cell cycle progression around the critical point through the PI3K-Akt pathway (Fig. 6D).

Actually, the identified critical transition as well as the enriched PI3K signaling pathway were supported by recent studies (*53,54*), in which the intricate interplay between the PI3K signaling pathway, encompassing PI3K and its downstream molecule Akt/PKB, and the preservation of self-renewal and multilineage differentiation potential in embryonic stem cells (ES cells) was found essential for cell proliferation and differentiation. This pathway has also exhibited significant involvement in orchestrating cell proliferation and differentiation across a spectrum of cell types (*70*). Figure 6D elucidates potential regulatory mechanisms for the functional analysis of stPCA-signaling genes and their immediate neighbors. Notably, growth factors such as extracellular matrix (ECM) components COL4A1 and SPP1 have emerged as influential upstream regulatory factors propelling cells through critical developmental transitions during the process of cell differentiation. These factors exhibit upregulation after critical transition, potentially steering downstream gene expressions. Furthermore, within the PI3K/Akt pathway, a responsive signaling cascade to stPCA-signaling genes has been discerned, exerting a pivotal influence on the regulation of cell proliferation. Upregulation of upstream stPCA-signaling genes COL4A1 and SPP1 initiates a cascade effect, culminating in the upregulation of downstream neighboring genes ITGB9, ITGB8, and PTK2. This, in turn, activates the AKT protein, leading to increased MDM2 gene expression and thereby fostering cell proliferation and differentiation. As a key stPCA-signaling gene, COL4A1 plays a critical role in cell adhesion, migration, proliferation, and differentiation through its interaction with integrin receptors. In summary, the induced downstream signaling reactions of stPCA-signaling genes actively engage in the intricate regulation of cell proliferation and differentiation.

Page 22

The sustained expression of stPCA-signaling genes in these pathways, both before and after critical transitions, highlights the pivotal role of identified critical transition in guiding embryonic stem cell development.

**Predicting the T2D transition risk based on continuous glucose monitoring data**

We applied stPCA to a continuous glucose monitoring (CGM) dataset (*71*) to analyze the risk of T2D. There were 208 patients selected from the outpatient clinic of hypertension and vascular risk of the University Hospital of Móstoles, in Madrid, from January 2012 to May 2015. A CGM record was obtained for 48 hours with sampling every 5 minutes. Patients were then followed every 6 months until the diagnosis of T2D or end of study. There were 17 confirmed cases of T2D, with a median time to diagnosis of 33.8 months after the sampling time point. For each subject, the CGM time-series data were segmented according to the following rules:

- Segmentation was performed based on the three main meals: breakfast, lunch, and dinner.
- Meal times were determined using local maxima in the CGM data. Breakfast: within the time range 6:00 to 9:00, identified by the local maximum T1, corresponding to the period [T1-3h, T1+1h]. Lunch: within the time range 11:00 to 14:00, identified by the local maximum T2, corresponding to the period [T2-3h, T2+1h]. Dinner: within the time range 17:00 to 21:00, identified by the local maximum T3, corresponding to the period [T3-3h, T3+1h].
- Additionally, two segments were added for the early morning (2:00 to 6:00) and fasting (21:00 to 1:00 of the next day) periods on the first day. Consequently, each subject's CGM time-series was divided into eight segments, each containing CGM data measured over a 4-hour interval, representing a 48-dimensional vector.

In this process, an 8-dimensional time series was generated for each case. Subsequently, we applied stPCA to this high-dimensional time series and computed the SD of the latent variable $z$. A substantial fluctuation (SD exceeding the average value) in $z$ indicates an elevated risk of transitioning from healthy state to T2D for one case. 17 cases confirmed T2D were demonstrated in Fig. 6E. Notably, stPCA identified 10 cases with a high risk of developing T2D in the future, surpassing predictions made by the traditional DNB approach. Therefore, stPCA is effective in providing the early warnings for T2D risk based on CGM data.

**Discussion**

The stPCA framework is proposed to represent the dynamical features of observed high-dimensional time-series data by only a single latent variable through a spatiotemporal transformation that converts high-dimensional spatial information into low-dimensional temporal information. This representative variable is acquired in an analytical and interpretable way by solving a unique characteristic equation, and theoretically can preserve the dynamical property of the original high-dimensional space. The ability of identifying the impending critical transitions for stPCA is derived from CSD theory, requiring only samples from before the tipping point, while change point detection methods, e.g., GSL (*72*) and stdCPD (*73*) generally require samples from both before and after the tipping point. In addition, the comparisons between the analytical solution from stPCA and an iterative numerical optimization algorithms such as sequential quadratic programming (SQP) (*38*) for a coupled Lorenz system and a multiple-node network show that the former yields a much more stable and accurate result (SI Appendix Fig. S6 and S10). Compared with deep learning-based representation learning method (*22–25*), stPCA achieves high efficiency with the time complexity of $O(nL)$.

Similar to the Koopman operator which can theoretically linearize high-dimensional nonlinear systems into an infinite linear space (without approximation) through spectral decomposition of infinite eigenfunctions, stPCA solves an optimization problem to represent high-dimensional systems with only one singular latent variable by linear embeddings (with approximation) in an analytical way with one eigenvalue/eigenvector. However, eigenfunctions of the Koopman operator $\mathcal{K}$ need to be sought (without a general method) (*26, 28, 30*). Representations has proven difficult to be obtained, and are often intractably complex or uninterpretable. DNN and DMD can be utilized for approximating the Koopman operator, but they face other challenges related to accuracy, training sample size, and time cost. As for stPCA, eigenvectors of matrix $H(X)$ can be relatively easy to obtain (as shown in the subsection of "The stPCA algorithm"), and the latent representation variable is derived from one dominant eigenvalue and its eigenvector. This property makes stPCA effective even for small sample size or short time-series data, as well as analytically tractable and computationally fast even for large datasets.

The applications of stPCA to both theoretical models and real-world datasets demonstrate its effectiveness and robustness in temporal data representation, tipping-point detection, and risk analysis of complex systems. First, implementing stPCA on coupled Lorenz models provided evidence that this method is effective in achieving efficient dimensionality reduction to an ultralow-dimensional space and exhibits resistance to noise. In addition, for ultrashort time-series cases, the extra spatial information can be transformed into temporal information of a latent variable, thereby further enhancing the robustness of stPCA to noise interference. This is demonstrated in Fig. 2, illustrating stPCA's ability to transform spatiotemporal information. Second, through the prognosis analysis of various cohort groups for ICU patients, the significant fluctuations of the latent single variable $z$ for each individual patient may indicate the critical/tipping condition. Thus, a scheme was proposed by combining the SD of $z$ and 2-5 diagnosis-related items to determine whether a patient should be discharged from the ICU at a specific time point, as demonstrated in the applications to MIMIC-III and MIMIC-IV datasets. Third, we compared this stPCA-based scheme with seven other dimensionality reduction methods, revealing that stPCA consistently outperforms its counterparts.

From the above analyses and applications, it is seen that stPCA possesses several noteworthy advantages: (i) stPCA excels in achieving ultralow-dimensionality reduction for dynamical systems without distortion. Its innovative use of a delay embedding strategy eliminates the need for truncation (approximation), a fundamental requirement in other dimensionality reduction methods. (ii) The transformation matrix $W$ and the single variable $z$ are derived through an analytical solution by solving a unique characteristic equation, which makes stPCA tractable and allows fast computation scalable to large datasets. (iii) stPCA efficiently represents the dynamics or temporal features of high-dimensional time-series data using only a single latent variable. This one-dimensional variable is roughly considered to represent the center manifold near the tipping point, enabling effective detection of the impending critical transitions in dynamical systems with codim-1 local bifurcation by examining the SD of this single variable.

There are still challenges and promising directions that motivate future work. First, there are still several limitations associated with the linear embedding $Z = WX$. Similar to PCA, which is mainly used for cross-sectional samples with the truncation of principal components as the approximation, stPCA is used for time-series samples but without truncation. Both methods adopt linear embedding due to its simple and explainable form, making them widely applicable in various fields. Such a linear embedding may not accurately characterize the nonlinear dynamics or may make inaccurate prediction of nonlinear dynamics. However, rather than to predict the overall dynamics, our method aims to identify critical transition dynamics which actually explores such inaccuracy (e.g., strong fluctuations). Specifically, when the prediction of a system by stPCA at a certain period or state, i.e., near a tipping point, becomes inaccurate due to its high nonlinearity or strong fluctuations during this period or state, the linear-embedding features in a one-dimensional space can be employed to detect this tipping point based on the CSD theory or our critical collective fluctuation principle. In other words, the detection of the tipping point can be actually quantified based on the inaccuracy (or fluctuation) of prediction.

Another drawback about stPCA is its sensitivity to the scaling of certain coordinates, which may affect the quality or stability of the results. Therefore, normalization of the original data is an essential step in the stPCA procedure. Nonetheless, despite this scaling concern, stPCA remains effective in identifying the tipping point using one single latent variable, in contrast to PCA which necessitates multiple representation variables. Actually, to mitigate this issue, we can also transform the information not directly from the original data or space, but from a feature space converted with a deep learning framework in our future work.

Despite these limitations, stPCA is proficient in detecting the critical transitions across diverse fields, when compared with some previous works in predicting critical behaviors(*74*). It provides a novel approach to dimensionality reduction for

Page 25

*Science Advances*

high-dimensional time-series data of nonlinear dynamical systems, characterized by computational efficiency, dynamics-preserving, accuracy, and robustness, thus holding significant promise for real-world applications.

**References**

1. M. Ledoux, *The Concentration of Measure Phenomenon* (American Mathematical Soc., 2001).

2. D. L. Donoho, High-dimensional data analysis: The curses and blessings of dimensionality. *AMS math challenges lecture* **1**, 32 (2000).

3. M. A. Carreira-Perpinán, A review of dimension reduction techniques. *Department of Computer Science. University of Sheffield. Tech. Rep. CS-96-09* **9**, 1–69 (1997).

4. D. Engel, L. Hüttenberger, B. Hamann, "A survey of dimension reduction methods for high-dimensional data analysis and visualization" in *Visualization of Large and Unstructured Data Sets: Applications in Geospatial Planning, Modeling and Engineering-Proceedings of IRTG 1131 Workshop 2011* (Schloss Dagstuhl-Leibniz-Zentrum fuer Informatik, 2012).

5. L. Chen, R. Liu, Z.-P. Liu, M. Li, K. Aihara, Detecting early-warning signals for sudden deterioration of complex diseases by dynamical network biomarkers. *Scientific reports* **2**, 1–8 (2012).

6. L. Van der Maaten, G. Hinton, Visualizing data using t-SNE. *Journal of machine learning research* **9** (2008).

7. E. Becht, L. McInnes, J. Healy, C.-A. Dutertre, I. W. Kwok, L. G. Ng, F. Ginhoux, E. W. Newell, Dimensionality reduction for visualizing single-cell data using UMAP. *Nature biotechnology* **37**, 38–44 (2019).

8. A. Buja, D. F. Swayne, M. L. Littman, N. Dean, H. Hofmann, L. Chen, Data visualization with multidimensional scaling. *Journal of computational and graphical statistics* **17**, 444–472 (2008).

9. S. T. Roweis, L. K. Saul, Nonlinear dimensionality reduction by locally linear embedding. *science* **290**, 2323–2326 (2000).

10. J. B. Tenenbaum, V. de Silva, J. C. Langford, A global geometric framework for nonlinear dimensionality reduction. *science* **290**, 2319–2323 (2000).

11. H. Abdi, L. J. Williams, Principal component analysis. *Wiley interdisciplinary reviews: computational statistics* **2**, 433–459 (2010).

12. A. J. Izenman, "Linear discriminant analysis" in *Modern Multivariate Statistical Techniques: Regression, Classification, and Manifold Learning* (Springer, 2013), pp. 237–280.

13. K. Kreutz-Delgado, J. F. Murray, B. D. Rao, K. Engan, T.-W. Lee, T. J. Sejnowski, Dictionary learning algorithms for sparse representation. *Neural computation* **15**, 349–396 (2003).

14. T.-W. Lee, "Independent Component Analysis" in *Independent Component Analysis* (Springer US, Boston, MA, 1998; http://link.springer.com/10.1007/978-1-4757-2851-4_2), pp. 27–66.

15. L. R. Medsker, L. Jain, Recurrent neural networks. *Design and Applications* **5**, 2 (2001).

16. F. Mastrogiuseppe, S. Ostojic, Linking connectivity, dynamics, and computations in low-rank recurrent neural networks. *Neuron* **99**, 609–623 (2018).

17. V. Thibeault, A. Allard, P. Desrosiers, The low-rank hypothesis of complex systems. *Nature Physics*, 1–9 (2024).

18. C. Lea, M. D. Flynn, R. Vidal, A. Reiter, G. D. Hager, "Temporal convolutional networks for action segmentation and detection" in *Proceedings of the IEEE Conference on Computer Vision and Pattern Recognition* (2017; http://openaccess.thecvf.com/content_cvpr_2017/html/Lea_Temporal_Convolutional_Networks_CVPR_2017_paper.html), pp. 156–165.

19. M. A. Kramer, Nonlinear principal component analysis using autoassociative neural networks. *AIChE journal* **37**, 233–243 (1991).

20. Y. Hua, J. Guo, H. Zhao, "Deep belief networks and deep learning" in *Proceedings of 2015 International Conference on Intelligent Computing and Internet of Things* (IEEE, 2015; https://ieeexplore.ieee.org/abstract/document/7111524/), pp. 1–4.

21. A. Vaswani, N. Shazeer, N. Parmar, J. Uszkoreit, L. Jones, A. N. Gomez, \Lukasz Kaiser, I. Polosukhin, Attention is all you need. *Advances in neural information processing systems* **30** (2017).

22. S. Schneider, J. H. Lee, M. W. Mathis, Learnable latent embeddings for joint behavioural and neural analysis. *Nature* **617**, 360–368 (2023).

23. C. Pandarinath, D. J. O'Shea, J. Collins, R. Jozefowicz, S. D. Stavisky, J. C. Kao, E. M. Trautmann, M. T. Kaufman, S. I. Ryu, L. R. Hochberg, Inferring single-trial neural population dynamics using sequential auto-encoders. *Nature methods* **15**, 805–815 (2018).

24. D. Floryan, M. D. Graham, Data-driven discovery of intrinsic dynamics. *Nature Machine Intelligence* **4**, 1113–1120 (2022).

25. A. Gosztolai, R. L. Peach, A. Arnaudon, M. Barahona, P. Vandergheynst, Interpretable statistical representations of neural population dynamics and geometry. arXiv arXiv:2304.03376 [Preprint] (2024). http://arxiv.org/abs/2304.03376.

26. I. Mezić, Spectral Properties of Dynamical Systems, Model Reduction and Decompositions. *Nonlinear Dyn* **41**, 309–325 (2005).

27. I. Mezić, Spectrum of the Koopman Operator, Spectral Expansions in Functional Spaces, and State-Space Geometry. *J Nonlinear Sci* **30**, 2091–2145 (2020).

28. S. L. Brunton, M. Budišić, E. Kaiser, J. N. Kutz, Modern Koopman Theory for Dynamical Systems. arXiv arXiv:2102.12086 [Preprint] (2021). http://arxiv.org/abs/2102.12086.

29. J. Axås, G. Haller, Model reduction for nonlinearizable dynamics via delay-embedded spectral submanifolds. *Nonlinear Dyn* **111**, 22079–22099 (2023).

30. B. Lusch, J. N. Kutz, S. L. Brunton, Deep learning for universal linear embeddings of nonlinear dynamics. *Nature communications* **9**, 4950 (2018).

31. S. L. Brunton, J. L. Proctor, J. N. Kutz, Discovering governing equations from data by sparse identification of nonlinear dynamical systems. *Proc. Natl. Acad. Sci. U.S.A.* **113**, 3932–3937 (2016).

32. F. Takens, "Detecting strange attractors in turbulence" in *Dynamical Systems and Turbulence, Warwick 1980* (Springer, 1981), pp. 366–381.

33. T. Sauer, J. A. Yorke, M. Casdagli, "Embedology," Journal of Statistical Physics. (1991).

34. H. Ma, S. Leng, K. Aihara, W. Lin, L. Chen, Randomly distributed embedding making short-term high-dimensional data predictable. *Proceedings of the National Academy of Sciences* **115**, E9994–E10002 (2018).

35. P. Chen, R. Liu, K. Aihara, L. Chen, Autoreservoir computing for multistep ahead prediction based on the spatiotemporal information transformation. *Nature communications* **11**, 1–15 (2020).

36. H. Peng, P. Chen, R. Liu, L. Chen, Spatiotemporal information conversion machine for time-series forecasting. *Fundamental Research* (2022).

37. P. Tao, X. Hao, J. Cheng, L. Chen, Predicting time series by data-driven spatiotemporal information transformation. *Information Sciences* **622**, 859–872 (2023).

38. P. T. Boggs, J. W. Tolle, Sequential quadratic programming. *Acta numerica* **4**, 1–51 (1995).

39. R. Liu, J. Wang, M. Ukai, K. Sewon, P. Chen, Y. Suzuki, H. Wang, K. Aihara, M. Okada-Hatakeyama, L. Chen, Hunt for the tipping point during endocrine resistance process in breast cancer by dynamic network biomarkers. *Journal of molecular cell biology* **11**, 649–664 (2019).

40. P. Chen, Y. Li, X. Liu, R. Liu, L. Chen, Detecting the tipping points in a three-state model of complex diseases by temporal differential networks. *Journal of Translational Medicine* **15**, 1–15 (2017).

41. Z. Zhong, J. Li, J. Zhong, Y. Huang, J. Hu, P. Zhang, B. Zhang, Y. Jin, W. Luo, R. Liu, MAPKAPK2, a potential dynamic network biomarker of α-synuclein prior to its aggregation in PD patients. *npj Parkinson's Disease* **9**, 41 (2023).

42. V. Dakos, M. Scheffer, E. H. Van Nes, V. Brovkin, V. Petoukhov, H. Held, Slowing down as an early warning signal for abrupt climate change. *Proceedings of the National Academy of Sciences* **105**, 14308–14312 (2008).

43. Y. Tong, R. Hong, Z. Zhang, K. Aihara, P. Chen, R. Liu, L. Chen, Earthquake alerting based on spatial geodetic data by spatiotemporal information transformation learning. *Proceedings of the National Academy of Sciences* **120**, e2302275120 (2023).

44. J. R. Tredicce, G. L. Lippi, P. Mandel, B. Charasse, A. Chevalier, B. Picqué, Critical slowing down at a bifurcation. *American Journal of Physics* **72**, 799–809 (2004).

45. R. Liu, X. Wang, K. Aihara, L. Chen, Early diagnosis of complex diseases by molecular biomarkers, network biomarkers, and dynamical network biomarkers. *Medicinal research reviews* **34**, 455–478 (2014).

46. J. H. Curry, A generalized Lorenz system. *Communications in Mathematical Physics* **60**, 193–204 (1978).

47. V. S. Anishchenko, A. N. Silchenko, I. A. Khovanov, Synchronization of switching processes in coupled Lorenz systems. *Physical Review E* **57**, 316 (1998).

48. L. Chen, R. Wang, C. Li, K. Aihara, *Modeling Biomolecular Networks in Cells: Structures and Dynamics* (Springer Science & Business Media, 2010).

49. A. Greenfield, A. Madar, H. Ostrer, R. Bonneau, DREAM4: Combining genetic and dynamic information to identify biological networks and dynamical models. *PloS one* **5**, e13397 (2010).

50. A. E. Johnson, T. J. Pollard, L. Shen, L. H. Lehman, M. Feng, M. Ghassemi, B. Moody, P. Szolovits, L. Anthony Celi, R. G. Mark, MIMIC-III, a freely accessible critical care database. *Scientific data* **3**, 1–9 (2016).

51. A. E. Johnson, L. Bulgarelli, L. Shen, A. Gayles, A. Shammout, S. Horng, T. J. Pollard, S. Hao, B. Moody, B. Gow, MIMIC-IV, a freely accessible electronic health record dataset. *Scientific data* **10**, 1 (2023).

52. E. R. Deyle, G. Sugihara, Generalized theorems for nonlinear state space reconstruction. *Plos one* **6**, e18295 (2011).

53. W. T. Lee, Tridiagonal matrices: Thomas algorithm. *MS6021, Scientific Computation, University of Limerick* (2011).

54. T. Eiter, H. Mannila, Computing discrete Fréchet distance. (1994).

55. P. K. Agarwal, R. B. Avraham, H. Kaplan, M. Sharir, Computing the discrete Fréchet distance in subquadratic time. *SIAM Journal on Computing* **43**, 429–449 (2014).

56. S. L. Brunton, B. W. Brunton, J. L. Proctor, E. Kaiser, J. N. Kutz, Chaos as an intermittently forced linear system. *Nature communications* **8**, 19 (2017).

57. M. Feng, J. I. McSparron, D. T. Kien, D. J. Stone, D. H. Roberts, R. M. Schwartzstein, A. Vieillard-Baron, L. A. Celi, Transthoracic echocardiography and mortality in sepsis: analysis of the MIMIC-III database. *Intensive care medicine* **44**, 884–892 (2018).

58. N. Hou, M. Li, L. He, B. Xie, L. Wang, R. Zhang, Y. Yu, X. Sun, Z. Pan, K. Wang, Predicting 30-days mortality for MIMIC-III patients with sepsis-3: a machine learning approach using XGboost. *Journal of translational medicine* **18**, 1–14 (2020).

59. J. Xu, B. S. Glicksberg, C. Su, P. Walker, J. Bian, F. Wang, Federated learning for healthcare informatics. *Journal of Healthcare Informatics Research* **5**, 1–19 (2021).

60. S. L. Hyland, M. Faltys, M. Hüser, X. Lyu, T. Gumbsch, C. Esteban, C. Bock, M. Horn, M. Moor, B. Rieck, Early prediction of circulatory failure in the intensive care unit using machine learning. *Nature medicine* **26**, 364–373 (2020).

61. L. Rasmy, Y. Xiang, Z. Xie, C. Tao, D. Zhi, Med-BERT: pretrained contextualized embeddings on large-scale structured electronic health records for disease prediction. *NPJ digital medicine* **4**, 86 (2021).

62. S. N. Golmaei, X. Luo, "DeepNote-GNN: predicting hospital readmission using clinical notes and patient network" in *Proceedings of the 12th ACM Conference on Bioinformatics, Computational Biology, and Health Informatics* (2021), pp. 1–9.

63. M. Müller, Dynamic time warping. *Information retrieval for music and motion*, 69–84 (2007).

64. D. J. Berndt, J. Clifford, "Using dynamic time warping to find patterns in time series" in *Proceedings of the 3rd International Conference on Knowledge Discovery and Data Mining* (1994), pp. 359–370.

65. F. Petitjean, A. Ketterlin, P. Gançarski, A global averaging method for dynamic time warping, with applications to clustering. *Pattern recognition* **44**, 678–693 (2011).

66. D. P. Kingma, M. Welling, An introduction to variational autoencoders. *Foundations and Trends® in Machine Learning* **12**, 307–392 (2019).

67. L.-F. Chu, N. Leng, J. Zhang, Z. Hou, D. Mamott, D. T. Vereide, J. Choi, C. Kendziorski, R. Stewart, J. A. Thomson, Single-cell RNA-seq reveals novel regulators of human embryonic stem cell differentiation to definitive endoderm. *Genome biology* **17**, 1–20 (2016).

68. J. Jia, L. Chen, Single-cell RNA sequencing data analysis based on non-uniform ε-neighborhood network. *Bioinformatics* **38**, 2459–2465 (2022).

69. L. Armstrong, O. Hughes, S. Yung, L. Hyslop, R. Stewart, I. Wappler, H. Peters, T. Walter, P. Stojkovic, J. Evans, The role of PI3K/AKT, MAPK/ERK and NFκβ signalling in the maintenance of human embryonic stem cell pluripotency and viability highlighted by transcriptional profiling and functional analysis. *Human molecular genetics* **15**, 1894–1913 (2006).

70. Y. Da, Y. Mou, M. Wang, X. Yuan, F. Yan, W. Lan, F. Zhang, Mechanical stress promotes biological functions of C2C12 myoblasts by activating PI3K/AKT/mTOR signaling pathway. *Mol Med Report*, doi: 10.3892/mmr.2019.10808 (2019).

71. C. Rodríguez De Castro, L. Vigil, B. Vargas, E. García Delgado, R. García Carretero, J. Ruiz-Galiana, M. Varela, Glucose time series complexity as a predictor of type 2 diabetes. *Diabetes Metabolism Res* **33**, e2831 (2017).

72. D. Sulem, H. Kenlay, M. Cucuringu, X. Dong, Graph similarity learning for change-point detection in dynamic networks. *Mach Learn* **113**, 1–44 (2024).

73. L. Li, J. Li, Online Change-Point Detection in High-Dimensional Covariance Structure with Application to Dynamic Networks. *Journal of Machine Learning Research* **24**, 1–44 (2023).

74. S. Tajima, T. Mita, D. J. Bakkum, H. Takahashi, T. Toyoizumi, Locally embedded presages of global network bursts. *Proc. Natl. Acad. Sci. U.S.A.* **114**, 9517–9522 (2017).

75. M. B. Kennel, R. Brown, H. D. I. Abarbanel, Determining embedding dimension for phase-space reconstruction using a geometrical construction. *Phys. Rev. A* **45**, 3403–3411 (1992).

## Acknowledgments

**Data and materials availability:** The simulation data of coupled Lorenz system and multiple-node-network and codes of the stPCA algorithm are available from https://github.com/RPcb/stPCA. The raw data of MIMIC-III and MIMIC-IV are available from https://physionet.org/content/mimiciii/1.4/ and https://physionet.org/content/mimiciv/2.2/. The continuous glucose monitoring (CGM) dataset is from https://doi.org/10.1371/journal.pone.0225817.s001. The dataset of single-cell embryonic development is deposited in NCBI's Gene Expression Omnibus and accessible through https://www.ncbi.nlm.nih.gov/geo/query/acc.cgi?acc=GSE75748. The signaling pathway analysis is carried out based on Kyoto Encyclopedia of Genes and Genomes (KEGG).

**Competing interests:** The authors declare that they have no conflict of interest.